\documentclass[11pt,a4paper]{article}
\usepackage[hyperref]{emnlp2020}

\usepackage[utf8x]{inputenc}

\usepackage{times}
\usepackage{graphicx}
\usepackage{subfigure}
\usepackage{natbib}
\usepackage{microtype}
\usepackage{amsmath}
\usepackage{amssymb}
\usepackage{multirow}
\usepackage{comment}
\usepackage{tabularx}
\usepackage{algpseudocode}
\usepackage{booktabs}
\usepackage{algorithm}

\usepackage{multicol}
\usepackage{pgfplotstable}
\pgfplotsset{compat=1.5}
\usepackage{tikz}
\usepackage{tabularx}
\usetikzlibrary{shapes, shadows,positioning}
\usepgfplotslibrary{fillbetween}
\pgfplotsset{
    tick label style = {font = \tiny},
    legend style = {font = \footnotesize},
}

\usepackage{amsmath,amsfonts,bm}

\def\eqref#1{equation~\ref{#1}}

\def\1{\bm{1}}

\def\vb{{\bm{b}}}
\def\vc{{\bm{c}}}

\def\ve{{\bm{e}}}

\def\vu{{\bm{u}}}

\def\vw{{\bm{w}}}
\def\vx{{\bm{x}}}

\def\vz{{\bm{z}}}
\def\vell{{\bm{\ell}}}
\def\vdelta{{\bm{\delta}}}

\def\mE{{\bm{E}}}
\def\mF{{\bm{F}}}

\def\mZ{{\bm{Z}}}

\DeclareMathAlphabet{\mathsfit}{\encodingdefault}{\sfdefault}{m}{sl}
\SetMathAlphabet{\mathsfit}{bold}{\encodingdefault}{\sfdefault}{bx}{n}

\def\gB{{\mathcal{B}}}

\def\gS{{\mathcal{S}}}

\def\gV{{\mathcal{V}}}

\def\gX{{\mathcal{X}}}

\newcommand{\R}{\mathbb{R}}

\newcommand{\softmax}{\mathrm{softmax}}

\DeclareMathOperator*{\argmax}{arg\,max}

\algtext*{EndWhile}
\algtext*{EndFor}
\algtext*{EndIf}
\algtext*{EndProcedure}
\algrenewcommand\algorithmicindent{1.0em}%
\mathchardef\mhyphen="2D 
\newcommand{\topk}{\mathrm{top\mhyphen k}}
\newcommand{\vphi}{\bm{\phi}}

\usepackage{xpatch}
\makeatletter
\xpatchcmd{\algorithmic}
  {\ALG@tlm\z@}{\leftmargin\z@\ALG@tlm\z@}
  {}{}
\makeatother

\makeatletter
\newcommand*{\algrule}[1][\algorithmicindent]{%
  \makebox[#1][l]{%
    \hspace*{.5em}
    \vrule height .75\baselineskip depth .25\baselineskip
  }
}

\newcount\ALG@printindent@tempcnta
\def\ALG@printindent{%
    \ifnum \theALG@nested>0
    \ifx\ALG@text\ALG@x@notext
    \else
    \unskip
    \ALG@printindent@tempcnta=1
    \loop
    \algrule[\csname ALG@ind@\the\ALG@printindent@tempcnta\endcsname]%
    \advance \ALG@printindent@tempcnta 1
    \ifnum \ALG@printindent@tempcnta<\numexpr\theALG@nested+1\relax
    \repeat
    \fi
    \fi
}
\patchcmd{\ALG@doentity}{\noindent\hskip\ALG@tlm}{\ALG@printindent}{}{\errmessage{failed to patch}}
\patchcmd{\ALG@doentity}{\item[]\nointerlineskip}{}{}{} 
\makeatother

\newcommand{\paragraphbe}[1]{\vspace{0.75ex}\noindent{\bf \em #1} }

\aclfinalcopy 
\def\aclpaperid{2005} 

\newcommand{\soft}[1]{\check{#1}}
\newcommand{\beam}{\vc_{1:t-1}}

\title{Adversarial Semantic Collisions}

\author{Congzheng Song \\
  Cornell University \\
  \texttt{cs2296@cornell.edu} \\\And
  Alexander M. Rush \\
  Cornell Tech \\
  \texttt{arush@cornell.edu} \\ \And
  Vitaly Shmatikov \\
  Cornell Tech \\
  \texttt{shmat@cs.cornell.edu} \\
}

\date{}

\begin{document}
\maketitle

\begin{abstract}

We study \emph{semantic collisions}: texts that are semantically unrelated
but judged as similar by NLP models.  We develop gradient-based approaches
for generating semantic collisions and demonstrate that state-of-the-art
models for many tasks which rely on analyzing the meaning and similarity
of texts\textemdash including paraphrase identification, document
retrieval, response suggestion, and extractive summarization\textemdash
are vulnerable to semantic collisions.  For example, given a target
query, inserting a crafted collision into an irrelevant document can
shift its retrieval rank from 1000 to top 3.  We show how to generate
semantic collisions that evade perplexity-based filtering and discuss
other potential mitigations.
Our code is available at \url{https://github.com/csong27/collision-bert}.
\end{abstract}

\section{Introduction}
\label{sec:introduction}


\begin{table*}[t!]
\centering
\footnotesize
\setlength{\tabcolsep}{5pt}
\begin{tabularx}{\textwidth}{c|>{\raggedright}X|c}
\toprule
Task &  Target inputs and collisions & Model output\\
\midrule
\multirow{5}{*}{\shortstack{Paraphrase \\ Identification}} & 
{\bf  \em Input $(\vx)$}: Does cannabis oil cure cancer? Or are the sellers hoaxing?
& \multirow{5}{*}{\shortstack{$\ge 99\%$ \\ confidence \\ of paraphrase}} \\
& {\bf  \em Aggressive $(\vc)$}: \textcolor{red}{\bf \_Pay 0ff your mortgage} der Seller chem Wad marijuana scarcity prince &  \\
& {\bf  \em Regularized aggressive $(\vc)$}: caches users remedies paved Sell Medical hey untold Caval OR and of of of of of of of of of of of of of of a a a of a & \\
& {\bf  \em Natural $(\vc)$}: he might actually work when those in & \\
 \midrule
\multirow{9}{*}{\shortstack{Document \\ Retrieval}} &
{\bf  \em Query $(\vx)$}: Health and Computer Terminals & \multirow{9}{*}{\shortstack{Irrelevant \\ articles' \\ ranks $\le$ 3}}\\
& {\bf  \em Aggressive $(\vc)$}: chesapeake oval mayo knuckles crowded double transmitter gig after nixon, tipped incumbent physician kai joshi astonished northwestern documents $|$ obliged dumont determines philadelphia consultative oracle keyboards dominates tel node  & \\
& {\bf \em Regularized aggressive $(\vc)$}: and acc near floors : panicked ; its employment became impossible, the – of cn magazine usa, in which " "'panic over unexpected noise, noise of and a of the of the of the of a of of the of the of of of of the of of of of the of of the of.  & \\ 
&{\bf \em Natural $(\vc)$}: the ansb and other buildings to carry people : three at the mall, an infirmary, an auditorium, and a library, as well as a clinic, pharmacy, and restaurant  & \\
\midrule
\multirow{5}{*}{\shortstack{Response \\ Suggestion}} &
{\bf \em Context $(\vx)$}: ...i went to school to be a vet , but i didn't like it.  &  \multirow{5}{*}{\shortstack{$\vc$'s rank = 1}} \\
& {\bf \em Aggressive $(\vc)$}:  \textcolor{red}{\bf buy v1agra in canadian pharmacy}  to breath as four ranger color &
\\
& {\bf \em Regularized aggressive $(\vc)$}: kill veterans and oxygen snarled clearly you were a a to to and a a to to to to to to to to to to  & 
\\
& {\bf \em Natural $(\vc)$}: then not have been an animal, or a human or a soldier but should &
\\
 \midrule
\multirow{6}{*}{\shortstack{Extractive \\ Summarization}} &
{\bf  \em Truth}: on average, britons manage just six and a half hours ' sleep a night , which is far less than the recommended eight hours.  &  \multirow{6}{*}{\shortstack{$\vc$'s rank = 1}}  \\
& {\bf  \em Aggressive $(\vc)$}: iec cu franks believe carbon chat fix pay carbon targets co₂ 8 iec cu mb &\\
& {\bf \em Regularized aggressive $(\vc)$}: the second mercury project carbon b mercury is a will produce 38 million 202 carbon a a to to to to to to to to to to to to to &\\
& {\bf \em Natural $(\vc)$}: 1 million men died during world war ii; over 40 percent were women  &\\
\bottomrule
\end{tabularx}
\caption{\footnotesize Four tasks in our study.  Given an input $\vx$ and white-box
access to a victim model, the adversary produces a collision $\vc$
resulting in a deceptive output.  Collisions can be nonsensical or
natural-looking and also carry spam messages (shown in red).}
\label{tab:example}
\end{table*}



Deep neural networks are vulnerable to adversarial
examples~\cite{szegedy2014intriguing,goodfellow2015explaining},
i.e., imperceptibly perturbed inputs that cause models to make
wrong predictions.  Adversarial examples based on inserting or
modifying characters and words have been demonstrated for text
classification~\cite{liang2018deep, ebrahimi2018hotflip, pal2020transfer},
question answering~\cite{jia2017adversarial, wallace2019universal}, and
machine translation~\cite{belinkov2018synthetic, wallace2020imitation}.
These attacks aim to minimally perturb the input so as it to preserve
its semantics while changing the output of the model.

In this work, we introduce and study a different class of vulnerabilities
in NLP models for analyzing the meaning and similarity of texts.
Given an input (query), we demonstrate how to generate a \textbf{semantic
collision}: an unrelated text that is judged semantically equivalent by
the target model.  Semantic collisions are the ``inverse'' of adversarial
examples.  Whereas adversarial examples are similar inputs that produce
dissimilar model outputs, semantic collisions are dissimilar inputs that
produce similar model outputs.


We develop gradient-based approaches for generating collisions given
white-box access to a model and deploy them against several NLP tasks.
For paraphrase identification, the adversary crafts collisions that are
judged as a valid paraphrase of the input query; downstream applications
such as removing duplicates or merging similar content will thus
erroneously merge the adversary's inputs with the victim's inputs.
For document retrieval, the adversary inserts collisions into one
of the documents that cause it to be ranked very high even though it
is irrelevant to the query.  For response suggestion, the adversary's
irrelevant text is ranked as the top suggestion and can also carry spam
or advertising.  For extractive summarization, the adversary inserts
a collision into the input text, causing it to be picked as the most
relevant content.

Our first technique generates collisions aggressively, without regard
to potential defenses.  We then develop two techniques, ``regularized
aggressive'' and ``natural,'' that constrain generated collisions using
a language model so as to evade perplexity-based filtering.  We evaluate
all techniques against state-of-the-art models and benchmark datasets on
all four tasks.  For paraphrase identification on Quora question pairs,
our collisions are (mis)identified as paraphrases of inputs with 97\%
confidence on average.  For document retrieval, our collisions shift the
median rank of irrelevant documents from 1000 to around 10.  For response
suggestion in dialogue (sentence retrieval), our collisions are ranked
as the top response 99\% and 86\% of the time with the aggressive and
natural techniques, respectively.  For extractive summarization, our
collisions are chosen by the model as the summary 100\% of the time. We conclude by discussing potential defenses against these attacks.

\section{Related Work}


\noindent
\textbf{\em Adversarial examples in NLP.}
Most of the previously studied adversarial attacks in NLP aim
to minimally modify or perturb inputs while changing the model's output.
\citet{hosseini2017deceiving} showed that perturbations, such as inserting dots or spaces between characters, can
deceive a toxic comment classifier.
HotFlip used gradients to find such perturbations 
given white-box access to the target model~\cite{ebrahimi2018hotflip}.  
\citet{wallace2019universal} extended HotFlip by inserting a short crafted ``trigger''
text to any input as perturbation; the trigger words are often highly associated with the target
class label.
Other approaches are based on rules, 
heuristics or generative models~\cite{mahler2017breaking, ribeiro2018semantically,iyyer2018adversarial, zhao2018generating}.
As explained in Section~\ref{sec:introduction},
our goal is the inverse of adversarial examples: we aim to generate inputs
with drastically different semantics that are perceived as similar by
the model.

Several works studied attacks that change the semantics of inputs.
\citet{jia2017adversarial} showed that inserting a heuristically
crafted sentence into a paragraph can trick a question answering
(QA) system into picking the answer from the inserted sentence.
Aggressively perturbed texts based on HotFlip are nonsensical and can be
translated into meaningful and malicious outputs by black-box translation
systems~\cite{wallace2020imitation}.  Our semantic collisions extend the
idea of changing input semantics to a different class of NLP models; we
design new gradient-based approaches that are not perturbation-based and
are more effective than HotFlip attacks; and, in addition to nonsensical
adversarial texts, we show how to generate ``natural'' collisions that
evade perplexity-based defenses.

\paragraphbe{Feature collisions in computer vision.}
Feature collisions have been studied in image analysis models.
\citet{jacobsen2019excessive} showed that images from different classes
can end up with identical representations due to excessive invariance of
deep models.  An adversary can modify the input to change its class while
leaving the model's prediction unaffected~\cite{jacobsen2019exploiting}.
The intrinsic property of rectifier activation function can
cause images with different labels to have the same feature
vectors~\cite{li2019approximate}.






\section{Threat Model}


We describe the targets of our attack, the threat model, and the
adversary's objectives.

\paragraphbe{Semantic similarity.}
\label{sec:semantic}
Evaluating semantic similarity of a pair of texts is at the core
of many NLP applications.  \emph{Paraphrase identification} decides
whether sentences are paraphrases of each other and can be used to
merge similar content and remove duplicates.  \emph{Document retrieval}
computes semantic similarity scores between the user's query and each
of the candidate documents and uses these scores to rank the documents.
\emph{Response suggestion}, aka Smart Reply~\cite{kannan2016smart}
or sentence retrieval, selects a response from a pool of candidates
based on their similarity scores to the user's input in dialogue.
\emph{Extractive summarization} ranks sentences in a document based
on their semantic similarity to the document's content and outputs the
top-ranked sentences.

For each of these tasks, let $f$ denote the model and $\vx_a, \vx_b$ a
pair of text inputs.  There are two common modeling approaches for these
applications.  In the first approach, the model takes the concatenation
$\oplus$ of $\vx_a$ and $\vx_b$ as input and directly produces a
similarity score $f(\vx_a \oplus \vx_b)$.  In the second approach,
the model computes a sentence-level embedding $f(\vx)\in\R^h$, i.e.,
a dense vector representation of input $\vx$.  The similarity score is
then computed as $s(f(\vx_a), f(\vx_b))$, where $s$ is a vector similarity
metric such as cosine similarity.  Models based on either approach are
trained with similar losses, such as the binary classification loss where
each pair of inputs is labeled as 1 if semantically related, 0 otherwise.
For generality, let $\gS(\cdot, \cdot)$ be a similarity function that
captures semantic relevance under either approach.  We also assume
that $f$ can take $\vx$ in the form of a sequence of discrete words
(denoted as $w$) or word embedding vectors (denoted as $\ve$), depending
on the scenario.


\paragraphbe{Assumptions.} 
We assume that the adversary has full knowledge of the target model,
including its architecture and parameters.  It may be possible
to transfer white-box attacks to the black-box scenario using model
extraction~\cite{krishna2019thieves,wallace2020imitation}; we leave this to future work.
The adversary controls some inputs that will be used by the target
model, e.g., he can insert or modify candidate documents for a
retrieval system.

\paragraphbe{Adversary's objectives.} 
Given a target model $f$ and target sentence $\vx$, the adversary
wants to generate a collision $\vx_b = \vc$ such that $f$ perceives
$\vx$ and $\vc$ as semantically similar or relevant.  Adversarial
uses of this attack depend on the application.  If an application
is using paraphrase identification to merge similar contents, e.g., in
Quora~\cite{quoramerge},
the adversary can use collisions to deliver spam or advertising to users.
In a retrieval system, the adversary can use collisions to boost the
rank of irrelevant candidates for certain queries.  For extractive
summarization, the adversary can cause collisions to be returned as the
summary of the target document.

\section{Adversarial Semantic Collisions}
\begin{figure}[t!]
\centering
\includegraphics[width=0.48\textwidth]{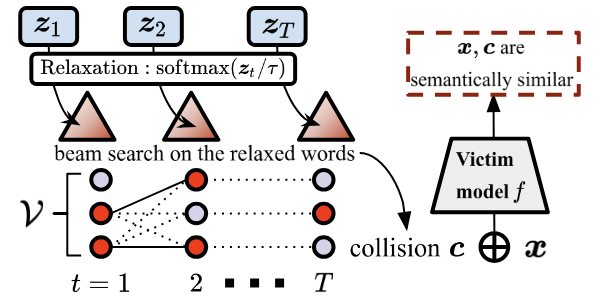}
\caption{\footnotesize Overview of generating semantic collision $\vc$
for a query input $\vx$.  The continuous variables $\vz_t$ relax the
words in $\vc$ and are optimized with gradients.  We search in the
simplex produced by $\vz_t$ for the actual colliding words in $\vc$.
}
\label{fig:my_label}
\end{figure}

\begin{algorithm*}
\caption{Generating adversarial semantic collisions}
\footnotesize
\textbf{Input:} input text  $\vx$,  similarity function $\gS$, embeddings $\mE$, language model $g$, vocabulary $\gV$, length $T$ \\
\textbf{Hyperparams:} beam size $B$, top-k size $K$, iterations $N$,  step size $\eta$, temperature $\tau$, score coefficient $\beta$,  label smoothing  $\epsilon$
\begin{multicols}{2}
\begin{algorithmic}
\Procedure{Main}{}
\State \Return collision $\vc=$\textsc{Aggressive}() or \textsc{Natural()} 
\EndProcedure
\Procedure{Aggressive}{}
\State $\mZ\gets[\vz_1, \ldots, \vz_T], \vz_t\gets\bm{0}\in\R^{|\gV|}$
\While{similarity score not converged}
\For{iteration 1 to $N$}
\State $\soft{\vc}\gets[\soft{\vc}_1,\ldots,\soft{\vc}_T], \soft{\vc}_t\gets\softmax(\vz_t / \tau)$   
\State $\mZ\gets\mZ + \eta\cdot\nabla_\mZ (1-\beta)\cdot\gS(\vx, \soft{\vc}) + \beta\cdot\Omega(\mZ)$ 
\EndFor
\State $\gB\gets B$ replicates of empty token 
\For{$t=1$ to $T$}
\State $\mF_t\gets\bm{0}\in \R^{B\times K}$, beam score matrix
\For{$\beam\in\gB, w\in\topk(\vz_t, K)$}   
\State $\mF_t[\beam, w]  \gets \gS(\vx, \beam\oplus w \oplus \soft{\vc}_{t+1:T})$ 
\EndFor
\State $\gB\gets \{\beam\oplus w| (\beam, w) \in\topk(\mF_t, B)\}$ 
\EndFor
\State $\mathrm{LS}(\vc_t) \gets$ Eq~\ref{eq:ls} with $\epsilon$ for $\vc_t\in \argmax \gB$ 
\State $\vz_t \gets \log\mathrm{LS}(\vc_t)$ for $\vz_t$ in $\mZ$ 
\EndWhile
\State \Return $\vc = \argmax \gB$
\EndProcedure
\Procedure{Natural}{}
\State $\gB\gets B$ replicates of start token 
\For{$t=1$ to $T$}
\State $\mF_t\gets\bm{0}\in \R^{B\times K}$, beam score matrix
\For{each beam $\beam\in \gB$}
\State $\vell_t \gets g(\beam)$, next token logits from LM
\State $\vz_t \gets$ \textsc{PerturbLogits}($\vell_t, \beam$) 
\For{$w\in \topk(\vz_t, K)$}    
\State $\mF_t[\beam, w]  \gets $ joint score from Eq~\ref{eq:local}
\EndFor
\State $\gB\gets \{\beam\oplus w| (\beam, w) \in \topk(\mF_t, B)\}$ 
\EndFor
\EndFor
\State \Return $\vc = \argmax \gB$
\EndProcedure
\Procedure{PerturbLogits}{$\vell, \beam$} 
\State $\vdelta \gets \bm{0}\in\R^{|\gV|}$  
\For{iteration 1 to $N$}
\State $\soft{\vc}_t \gets \softmax((\vell + \vdelta) / \tau)$
\State $\vdelta \gets \vdelta + \eta \cdot \nabla_{\vdelta}\gS(\vx, \beam\oplus \soft{\vc}_t)$  
\EndFor
\State \Return $\vz=\vell + \vdelta$
\EndProcedure
\end{algorithmic}
\end{multicols}
\label{alg:collision}
\end{algorithm*}

Given an input (query) sentence $\vx$, we aim to generate a
collision $\vc$ for the victim model with the white-box similarity
function $\gS$.  This can be formulated as an optimization problem:
$\arg\max_{\vc\in\gX} \gS(\vx, \vc)$ such that $\vx$ and $\vc$
are semantically unrelated.  A brute-force enumeration of $\gX$ is
computationally infeasible.  Instead, we design gradient-based approaches
outlined in Algorithm~\ref{alg:collision}.  We consider two variants:
(a) aggressively generating unconstrained, nonsensical collisions, and
(b) constrained collisions, i.e., sequences of tokens that appear fluent
under a language model and cannot be automatically filtered out based
on their perplexity.

We assume that models can accept inputs as both hard one-hot words
and soft words,\footnote{For a soft-word input, models will compute
the word vector as the weighted average of word embeddings by the
probability vector.} where a soft word is a probability vector
$\soft{\vw}\in\Delta^{|\gV|-1}$ for vocabulary $\gV$.

\subsection{Aggressive Collisions}
\label{sec:nonsensical}

We use gradient-based search to generate a fixed-length collision given
a target input.  The search is done in two steps: 1) we find a continuous
representation of a collision using gradient optimization with relaxation,
and 2) we  apply beam search to produce a hard collision.  We repeat
these two steps iteratively until the similarity score $\gS$ converges.


\paragraphbe{Optimizing for soft collision.} 
We first relax the optimization to a continuous representation with
temperature annealing.
Given the model's vocabulary $\gV$ and a fixed length $T$, we model word
selection at each position $t$ as a continuous logit vector
$\vz_t\in\R^{|\gV|}$.
To convert each $\vz_t$ to an input word, we model a softly
selected word at $t$ as:
\begin{align}
\label{eq:relax}
    \soft{\vc}_t = \softmax(\vz_t / \tau)
\end{align}
where $\tau$ is a temperature scalar.
Intuitively, softmax on $\vz_t$
gives the probability of each word in $\gV$.  The temperature controls
the sharpness of word selection probability; when $\tau\rightarrow 0$,
the soft word $\soft{\vc}_t$ is the same as the hard word $\argmax \vz_t$.

We optimize for the continuous values $\vz$.
At each step, the soft word collisions
$\soft{\vc}=[\soft{\vc}_1,\ldots,\soft{\vc}_T]$ are forwarded to $f$ to
calculate $\gS(\vx, \soft{\vc})$.  Since all operations are continuous,
the error can be back-propagated all the way to each $\vz_t$ to calculate
its gradients.  We can thus apply gradient ascent to improve 
the objective.

\paragraphbe{Searching for hard collision.}  
After the relaxed optimization, we apply a projection step to find a
hard collision using discrete search.\footnote{We could project the soft
collision by annealing the temperature to $0$, $\vc = [\argmax\vz_1,
\dots, \argmax\vz_T].$ However, this approach yields sub-optimal results
because the hard $\argmax$ discards information from nearby words.}
Specifically, we apply left-to-right beam search on each $\vz_t$.
At every search step $t$, we first get the top $K$ words $w$ based on
$\vz_t$ and rank them by the target similarity
$\gS(\vx, \beam\oplus w \oplus \soft{\vc}_{t+1:T})$, where
$\soft{\vc}_{t+1:T}$ is the partial soft collision starting at $t+1$.
This procedure allows us to find a hard-word replacement for the soft
word at each position $t$ based on the previously found hard words and
relaxed estimates of future words.

\paragraphbe{Repeating optimization with hard collision.}
If the similarity score still has room for improvement after the beam
search, we use the current $\vc$ to initialize the soft solution $\vz_t$
for the next iteration of optimization by transferring the hard solution
back to continuous space.

In order to initialize the continuous relaxation from a hard sentence,
we apply label smoothing $(\mathrm{LS})$ to its one-hot representation.
For each word $\vc_t$ in the current $\vc$, we soften its one-hot vector
to be inside $\Delta^{|\gV| - 1}$ with
\begin{align}
\label{eq:ls}
\mathrm{LS}(\vc_t)_w =     
\begin{cases}
  1 - \epsilon & \text{if $w = \argmax\vc_t$}\\
  \frac{\epsilon}{|\gV| - 1} & \text{otherwise}
\end{cases}
\end{align}
where $\epsilon$ is
the label-smoothing parameter.
Since $\mathrm{LS}(\vc_t)$ is constrained in the probability simplex
$\Delta^{|\gV| - 1}$, we set each $\vz_t$ to $\log\mathrm{LS}(\vc_t)\in
\R^{|\gV|}$ as the initialization for optimizing the soft solution in
the next iteration.

\subsection{Constrained Collisions}


The Aggressive approach is very effective at finding collisions, but it can 
output nonsensical sentences. 
Since these sentences have high perplexity under a
language model (LM), simple filtering can eliminate them from consideration.
To evade perplexity-based filtering, we impose a soft constraint on collision
generation and jointly maximize target similarity and LM likelihood:
\begin{align}
\label{eq:constrained}
\max_{\vc\in\gX} (1 - \beta) \cdot \gS(\vx, \vc) + \beta \cdot \log P(\vc ;g) 
\end{align}
where $P(\vc ;g)$ is the LM likelihood for collision $\vc$ under a
pre-trained LM $g$ and $\beta\in[0,1]$ is an interpolation coefficient. 

We investigate two different approaches for solving the optimization
in equation~\ref{eq:constrained}: (a) adding a regularization term on
soft $\soft{\vc}$ to approximate the LM likelihood, and (b) steering a
pre-trained LM to generate natural-looking $\vc$.

\subsubsection{Regularized Aggressive Collisions}
Given a language model $g$, we can incorporate a soft version 
of the LM likelihood as a regularization term on the
soft aggressive $\soft{\vc}$ computed from the variables $ [\vz_1, \ldots, \vz_T]$:
\begin{align}
\label{eq:omega}
\Omega = \sum_{t=1}^T H(\soft{\vc}_t, P(w_t|\soft{\vc}_{1:t-1};g))
\end{align}
where $H(\cdot,\cdot)$ is cross entropy, $P(w_t|\soft{\vc}_{1:t-1};g)$
are the next-token prediction probabilities at $t$ given partial soft
collision $\soft{\vc}_{1:t-1}$.  Equation~\ref{eq:omega} relaxes the
LM likelihood on hard collisions by using soft collisions as input,
and can be added to the objective function for gradient optimization.
The variables $\vz_t$ after optimization will favor words that maximize
the LM likelihood.

To further reduce the perplexity of $\vc$, we exploit the degeneration
property of LM, i.e., the observation that LM assigns low perplexity
to repeating common tokens~\cite{holtzman2020curious}, and constrain a
span of consecutive tokens in $\vc$ (e.g., second half of $\vc$) to be
selected from most frequent English words instead of the entire $\gV$.
This modification produces even more disfluent collisions, but they
evade LM-based filtering.

\subsubsection{Natural Collisions}
Our final approach aims to produce fluent, low-perplexity outputs. Instead
of relaxing and then searching, we search and then relax each step for
equation~\ref{eq:constrained}. This lets us integrate a hard language
model while selecting next words in continuous space.
In each step $t$, we maximize:
\begin{align}
\max_{w\in\gV} \quad &(1 - \beta)\cdot\gS(\vx, \beam\oplus w) + \nonumber \\ &\beta\cdot \log P(\beam\oplus w; g) \label{eq:local}
\end{align}
where $\beam$ is the beam solution found before $t$.  This sequential
optimization is essentially LM decoding with a joint search on the LM
likelihood and target similarity $\gS$, of the collision prefix.


Optimizing equation~\ref{eq:local} exactly requires ranking each $w\in\gV$
based on LM likelihood $\log P(\beam\oplus w;g)$ and similarity $\gS(\vx,
\beam\oplus w)$.  Evaluating LM likelihood for every word at each step
is efficient because we can cache $\log P(\beam;g)$ and compute the 
next-word probability in the standard manner.
However, evaluating an arbitrary similarity function $\gS(\vx, \beam\oplus w), \forall
w\in\gV$, requires $|\gV|$ forwarded passes to $f$, which can
be computationally expensive.

\paragraphbe{Perturbing LM logits.} 
Inspired by Plug and Play LM~\cite{dathathri2020plug}, we modify the LM
logits to take similarity into account.  We first let  $\vell_t=g(\beam)$
be the next-token logits produced by LM $g$ at step $t$. We then
optimize from this initialization to find an update that favors words
maximizing similarity.  Specifically, we let $\vz_t=\vell_t + \vdelta_t$
where $\vdelta_t \in\R^{|\gV|}$ is a perturbation vector. We then take
a small number of gradient steps on the relaxed similarity objective
$\max_{\vdelta_t}\gS(\vx, \beam \oplus \soft{\vc}_t)$ where $\soft{\vc}_t$
is the relaxed soft word as in equation~\ref{eq:relax}.
This encourages the next-word prediction distribution from the perturbed
logits, $\soft{\vc}_t$, to favor words that are likely to collide with
the input $\vx$.



\paragraphbe{Joint beam search.} 
After perturbation at each step $t$, we find the top $K$ most likely
words in $\soft{\vc}_t$.  This allows us to only evaluate $\gS(\vx,
\beam\oplus w)$ for this subset of words $w$ that are likely under the LM
given the current beam context.  We rank these top $K$ words based on the
interpolation of target loss and LM log likelihood.  We assign a score
to each beam $\vb$ and each top $K$ word as in equation~\ref{eq:local},
and update the beams with the top-scored words.

This process leads to a natural-looking decoded sequence because each
step utilizes the true words as input.  As we build up a sequence,
the search at each step is guided by the joint score of two objectives,
semantic similarity and fluency.


\section{Experiments}

\noindent
\textbf{\em Baseline.} 
We use a simple greedy baseline based on
HotFlip~\cite{ebrahimi2018hotflip}.  We initialize the collision text
with a sequence of repeating words, e.g., ``the'', and iteratively replace
all words.  In each iteration, we look at every position $t$ and flip the
current $w_t$ to $v$ that maximizes the first-order Taylor approximation
of target similarity $\gS$:
\begin{align}
\label{eq:hotflip}
    \argmax_{1 \le t\le T, v\in\gV} (\ve_i - \ve_v)^\top \nabla_{\ve_t} \gS(\vx, \vc)
\end{align}
where $\ve_t,\ve_v$ are the word vectors for $w_t$ and $v$.
Following prior HotFlip-based attacks~\cite{michel2019evaluation,
wallace2019universal, wallace2020imitation}, we evaluate $\gS$ using
the top $K$ words from Equation~\ref{eq:hotflip} and flip to the word
with the lowest loss to counter the local approximation.

\paragraphbe{LM for natural collisions.} 
For generating natural collisions, we need a LM $g$ that
shares the vocabulary with the target model $f$.  When targeting models
that do not share the vocabulary with an available LM, we fine-tune
another BERT with an autoregressive LM task on the Wikitext-103
dataset~\cite{merity2017pointer}.  When targeting models based on RoBERTa,
we use pretrained GPT-2~\cite{radford2019language} as the LM since the
vocabulary is shared.

\paragraphbe{Unrelatedness.}
To ensure that collisions $\vc$ are not semantically similar to inputs
$\vx$, we filter out words that are relevant to $\vx$ from $\gV$ when
generating $\vc$.  First, we discard non-stop words in $\vx$; then, we
discard 500 to 2,000 words in $\gV$ with the highest similarity score
$\gS(\vx, w)$.

\paragraphbe{Hyperparameters.}
We use Adam~\cite{kingma2015adam} for gradient ascent.  Detailed
hyperparameter setup can be found in table~\ref{tab:params} in
Appendix~\ref{sec:appendix}.

\paragraphbe{Notation.} 
In the following sections, we abbreviate HotFlip baseline as
\textbf{HF}; aggressive collisions as \textbf{Aggr.}; regularized
aggressive collisions as \textbf{Aggr.} $\boldsymbol{\Omega}$  where
$\Omega$ is the regularization term in equation~\ref{eq:omega}; and
natural collisions as \textbf{Nat.}

\subsection{Tasks and Models}

We evaluate our attacks on paraphrase identification, document
retrieval, response suggestions and extractive summarization.  Our
models for these applications are pretrained transformers, including
BERT~\cite{devlin2019bert} and RoBERTa~\cite{liu2019roberta}, fine-tuned
on the corresponding task datasets and matching state-of-the-art
performance.

\paragraphbe{Paraphrase detection.}
We use the Microsoft Research Paraphrase Corpus
(MRPC)~\cite{dolan2005automatically} and Quora Question Pairs
(QQP)~\cite{WinNT}, and attack the first 1,000 paraphrase pairs from
the validation set.

We target the BERT and RoBERTa base models for MRPC and QQP, respectively.
The models take in concatenated inputs $\vx_a, \vx_b$ and output
the similarity score as $\gS(\vx_a, \vx_b) = \mathrm{sigmoid}(f(\vx_a
\oplus \vx_b))$.  We fine-tune them with the suggested hyper-parameters.
BERT achieves 87.51\% F1 score on MRPC and RoBERTa achieves 91.6\%
accuracy on QQP, consistent with prior work.

\paragraphbe{Document retrieval.}
We use the Common Core Tracks from 2017 and 2018 (Core17/18).  They have
50 topics as queries and use articles from the New York Times Annotated
Corpus and TREC Washington Post Corpus, respectively.


Our target model is Birch~\cite{yilmaz2019applying,yilmaz2019cross}.
Birch retrieves 1,000 candidate documents using the BM25 and RM3
baseline~\cite{abdul2004umass} and re-ranks them using the similarity
scores from a fine-tuned BERT model.  Given a query $\vx_q$ and a document
$\vx_d$, the BERT model assigns similarity scores $\gS(\vx_q, \vx_i)$
for each sentence $\vx_i$ in $\vx_d$.  The final score used by Birch for
re-reranking is: $\gamma\cdot\gS_\textrm{BM25} + (1-\gamma)\cdot\sum_i
\kappa_i\cdot\gS(\vx_q, \vx_i)$ where $\gS_\textrm{BM25}$ is the baseline
BM25 score and $\gamma, \kappa_i$ are weight coefficients.  We use the
published models\footnote{\url{https://github.com/castorini/birch}}
and coefficient values for evaluation.

We attack similarity scores $\gS(\vx_q, \vx_i)$ by inserting sentences
that collide with $\vx_q$ into irrelevant $\vx_d$.  We filter out query
words when generating collisions $\vc$ so that term frequencies of
query words in $\vc$ are 0, thus inserting collisions does not affect
the original $\gS_\textrm{BM25}$.  For each of the 50 query topics,
we select irrelevant articles that are ranked from 900 to 1000 by Birch
and insert our collisions into these articles to boost their ranks.

\begin{table*}[t]
\centering
\footnotesize
\setlength{\tabcolsep}{3.75pt}
\begin{tabular}{l|rr|rr|rrr|rr|rr|rrr}
\toprule
$\vc$ type & \multicolumn{2}{c|}{MRPC} & \multicolumn{2}{c|}{QQP} & \multicolumn{3}{c|}{Core17/18} & \multicolumn{2}{c|}{Chat-Bi} & \multicolumn{2}{c|}{Chat-Poly} & \multicolumn{3}{c}{CNNDM} \\
& $\gS$ & \% Succ & $\gS$ & \% Succ & $\gS$ & $r\le 10$ & $\le 100$ & $\gS$ & $r = 1$ & $\gS$  & $r=1$ & $\gS$  & $r=1$ & $r\le 3$ \\  
\midrule
Gold  & 0.87 & - & 0.90 & - & 1.34 & - & - & 17.14 & - & 25.30 & - & 0.51 & - & - \\ 
\midrule
HF & 0.60 & 67.3\% & 0.55 &  54.8\% & -0.96 & 0.0\% & 16.5\% & 21.20 & 78.5\% & 28.82 & 73.1\% & 0.50 & 67.9\% & 96.5\% \\
Aggr. & 0.93 & 97.8\% & 0.98 & 97.3\% & 1.62 & 49.9\% & 86.7\% & 23.79 & 99.8\% & 31.94 & 99.4\% & 0.69 & 99.4\% & 100.0\%\\
Aggr. $\Omega$ & 0.69 & 81.0\% & 0.91 & 91.1\% & 0.86 & 20.6\% & 69.7\% & 21.66 & 92.9\% & 29.51 & 90.7\% & 0.58 & 90.7\% & 100.0\% \\ 
Nat. & 0.78 & 98.6\% & 0.88 & 88.8\% & 0.77 & 12.3\% & 60.6\% & 22.15 & 86.0\% & 31.10 & 86.6\% & 0.37 & 30.4\% & 77.7\% \\
\bottomrule
\end{tabular}
\caption{\footnotesize Attack results. $r$ is the rank of collisions among candidates. Gold denotes the ground truth.}
\label{tab:attack}
\end{table*}

\paragraphbe{Response suggestion.}
We use the Persona-chat (Chat) dataset of
dialogues~\cite{zhang2018personalizing}.  The task is to pick the
correct utterance in each dialogue context from 20 choices.
We attack the first 1,000 contexts from the validation set.

We use transformer-based Bi- and Poly-encoders that achieved
state-of-the-art results on this dataset~\cite{humeau2020poly}.
Bi-encoders compute a similarity score for the dialogue context
$\vx_a$ and each possible next utterance $\vx_b$ as $\gS(\vx_a,
\vx_b) = f_\textrm{pool}(\vx_a)^\top f_\textrm{pool}(\vx_b)$
where $f_\textrm{pool}(\vx)\in\R^{h}$ is the pooling-over-time
representation from transformers.  Poly-encoders extend Bi-encoders
compute $\gS(\vx_a, \vx_b) = \sum_{i=1}^T \alpha_i\cdot  f(\vx_a)_i^\top
f_\textrm{pool}(\vx_b)$ where $\alpha_i$ is the weight from attention
and $f(\vx_a)_i$ is the $i$th token's contextualized representation.
We use the published models\footnote{\url{https://parl.ai/docs/zoo.html}}
for evaluation.

\paragraphbe{Extractive summarization.}
We use the CNN / DailyMail (CNNDM) dataset~\cite{hermann2015teaching},
which consists of news articles and labeled overview highlights.
We attack the first 1,000 articles from the validation set.

Our target model is PreSumm~\cite{liu2019text}.  Given a text $\vx_d$,
PreSumm first obtains a vector representation $\vphi_i\in\R^h$ for
each sentence $\vx_i$ using BERT, and scores each sentence $\vx_i$
in the text as $\gS(\vx_d, \vx_i) = \mathrm{sigmoid}(\vu^\top
f(\vphi_1, \ldots,\vphi_T)_i)$ where $\vu$ is a weight vector,
$f$ is a sentence-level transformer, and $f(\cdot)_i$ is the
$i$th sentence's contextualized representation.  Our objective
is to insert a collision $\vc$ into $\vx_d$ such that the rank of
$\gS(\vx_d, \vc)$ among all sentences is high.  We use the published
models\footnote{\url{https://github.com/nlpyang/PreSumm}} for evaluation.

\subsection{Attack Results}

For all attacks, we report the similarity score $\gS$ between $\vx$
and $\vc$; the ``gold'' baseline is the similarity between $\vx$ and the
ground truth.  For MRPC, QQP, Chat, and CNNDM, the ground truth is the
annotated label sentences (e.g., paraphrases or summaries); for Core17/18,
we use the sentences with the highest similarity $\gS$ to the query.
For MRPC and QQP, we also report the percentage of successful collisions
with $\gS > 0.5$.  For Core17/18, we report the percentage of irrelevant
articles ranking in the top-10 and top-100 after inserting collisions.
For Chat, we report the percentage of collisions achieving top-1 rank.
For CNNDM, we report the percentage of collisions with the top-1 and
top-3 ranks (likely to be selected as summary).  Table~\ref{tab:attack}
shows the results.

On MRPC, aggressive and natural collisions achieve around 98\% success;
aggressive ones have higher similarity $\gS$.  With regularization
$\Omega$, success rate drops to 81\%.  On QQP, aggressive collisions
achieve 97\% vs.\ 90\% for constrained collisions.

On Core17/18, aggressive collisions shift the rank of almost half of the
irrelevant articles into the top 10.  Regularized and natural collisions
are less effective, but more than 60\% are still ranked in the top 100.
Note that query topics are compact phrases with narrow semantics, thus
it might be harder to find constrained collisions for them.

On Chat, aggressive collisions achieve rank of 1 more than 99\% of the
time for both Bi- and Poly-encoders.  With regularization $\Omega$,
success drops slightly to above 90\%.  Natural collisions are less
successful, with 86\% ranked as 1.

On CNNDM, aggressive collisions are almost always ranked as the top
summarizing sentence.  HotFlip and regularized collisions are in the
top 3 more than 96\% of the time.  Natural collisions perform worse,
with 77\% ranked in the top 3.

Aggressive collisions always beat HotFlip on all tasks; constrained
collisions are often better, too.  The similarity scores $\gS$ for
aggressive collisions are always higher than for the ground truth.

\begin{table}[t]
\centering
\footnotesize
\setlength{\tabcolsep}{5pt}
\begin{tabular}{l|rrrrr}
\toprule
$\vc$ type & MRPC & QQP & Core & Chat & CNNDM \\
& $F_\text{BERT}$ & $F_\text{BERT}$ & $P_\text{BERT}$ & $P_\text{BERT}$ & $F_\text{BERT}$ \\ 
\midrule
Gold  & 0.66 & 0.68 & 0.17 & 0.14 & 0.38 \\ 
\midrule
Aggr.  & -0.22 & -0.17 & -0.34 & -0.31 & -0.31   \\
Aggr. $\Omega$ & -0.34 & -0.34 & -0.48  & -0.43 & -0.36 \\ 
Nat.  & -0.12 & -0.09 & -0.11 & -0.10 & -0.25 \\ 
\bottomrule
\end{tabular}
\caption{\footnotesize \textsc{BERTScore} between collisions and target inputs.
Gold denotes the ground truth.}
\label{tab:bertscore}
\end{table}

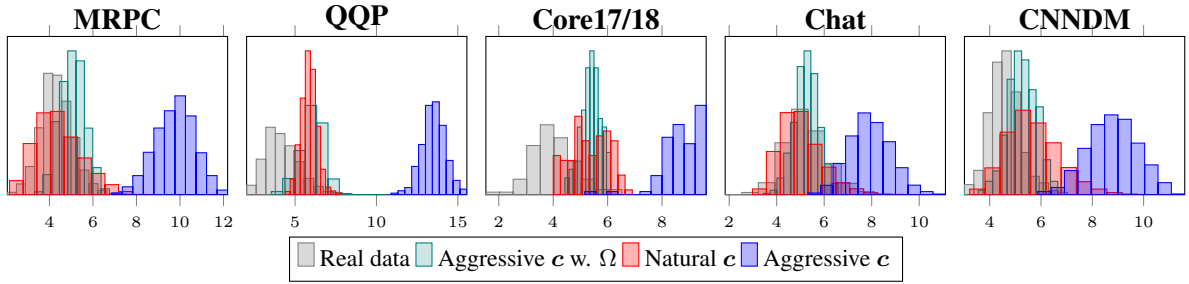
\begin{figure*}[t!]
    \centering
\begin{tikzpicture}
\begin{axis}[ybar, ytick=\empty, width=0.28\textwidth, height=0.23\textwidth, enlarge x limits=false, ymin=0., title style={yshift=-1.5ex}, title=\textbf{MRPC}, legend style={at={(0.01,0.8)}, anchor=west},
name=ax1]
\pgfplotstableread{plot/mrpc_perp.txt}\mydata;
\addplot[ybar interval, draw=gray, fill=gray, fill opacity=0.3] 
table [x expr=\thisrow{bt},y expr=\thisrow{nt}] \mydata;
\addplot[ybar interval, draw=teal, fill=teal, fill opacity=0.2] 
table [x expr=\thisrow{br},y expr=\thisrow{nr}] \mydata;
\addplot[ybar interval, draw=red, fill=red, fill opacity=0.3] 
table [x expr=\thisrow{ba},y expr=\thisrow{na}] \mydata;
\addplot[ybar interval, draw=blue, fill=blue, fill opacity=0.3] 
table [x expr=\thisrow{bn},y expr=\thisrow{nn}] \mydata;
\end{axis}
\begin{axis}[ybar, ytick=\empty, width=0.28\textwidth, height=0.23\textwidth, enlarge x limits=false, ymin=0., title style={yshift=-1.5ex}, title=\textbf{QQP}, legend style={at={(0.01,0.8)}, anchor=west},
name=ax2, at={(ax1.south east)}, xshift=0.25cm]
\pgfplotstableread{plot/qqp_perp.txt}\mydata;
\addplot[ybar interval, draw=gray, fill=gray, fill opacity=0.3] 
table [x expr=\thisrow{bt},y expr=\thisrow{nt}] \mydata;
\addplot[ybar interval, draw=teal, fill=teal, fill opacity=0.2] 
table [x expr=\thisrow{br},y expr=\thisrow{nr}] \mydata;
\addplot[ybar interval, draw=red, fill=red, fill opacity=0.3] 
table [x expr=\thisrow{ba},y expr=\thisrow{na}] \mydata;
\addplot[ybar interval, draw=blue, fill=blue, fill opacity=0.3] 
table [x expr=\thisrow{bn},y expr=\thisrow{nn}] \mydata;
\end{axis}
\begin{axis}[ybar, ytick=\empty, width=0.28\textwidth, height=0.23\textwidth, enlarge x limits=false, ymin=0., title style={yshift=-1.5ex}, title=\textbf{Core17/18}, legend style={at={(0.5, -0.55)},  anchor=south, legend columns=4},
name=ax3, at={(ax2.south east)}, xshift=0.25cm]
\pgfplotstableread{plot/doc_retr_perp.txt}\mydata;
\addplot[ybar interval, draw=gray, fill=gray, fill opacity=0.3] 
table [x expr=\thisrow{bt},y expr=\thisrow{nt}] \mydata;
\addplot[ybar interval, draw=teal, fill=teal, fill opacity=0.2] 
table [x expr=\thisrow{br},y expr=\thisrow{nr}] \mydata;
\addplot[ybar interval, draw=red, fill=red, fill opacity=0.3] 
table [x expr=\thisrow{ba},y expr=\thisrow{na}] \mydata;
\addplot[ybar interval, draw=blue, fill=blue, fill opacity=0.3] 
table [x expr=\thisrow{bn},y expr=\thisrow{nn}] \mydata;
\legend{Real data, Aggressive $\vc$ w. $\Omega$, Natural $\vc$, Aggressive $\vc$}
\end{axis}
\begin{axis}[ybar, ytick=\empty, width=0.28\textwidth, height=0.23\textwidth, enlarge x limits=false, ymin=0., title style={yshift=-1.5ex}, title=\textbf{Chat}, legend style={at={(0.01,0.8)}, anchor=west},
name=ax5, at={(ax3.south east)}, xshift=0.25cm]
\pgfplotstableread{plot/poly_perp.txt}\mydata;
\addplot[ybar interval, draw=gray, fill=gray, fill opacity=0.3] 
table [x expr=\thisrow{bt},y expr=\thisrow{nt}] \mydata;
\addplot[ybar interval, draw=teal, fill=teal, fill opacity=0.2] 
table [x expr=\thisrow{br},y expr=\thisrow{nr}] \mydata;
\addplot[ybar interval, draw=red, fill=red, fill opacity=0.3] 
table [x expr=\thisrow{ba},y expr=\thisrow{na}] \mydata;
\addplot[ybar interval, draw=blue, fill=blue, fill opacity=0.3] 
table [x expr=\thisrow{bn},y expr=\thisrow{nn}] \mydata;
\end{axis}
\begin{axis}[ybar, ytick=\empty, width=0.28\textwidth, height=0.23\textwidth, enlarge x limits=false, ymin=0., title style={yshift=-1.5ex}, title=\textbf{CNNDM},
name=ax6, at={(ax5.south east)}, xshift=0.25cm]
\pgfplotstableread{plot/ext_sum_perp.txt}\mydata;
\addplot[ybar interval, draw=gray, fill=gray, fill opacity=0.3] 
table [x expr=\thisrow{bt},y expr=\thisrow{nt}] \mydata;
\addplot[ybar interval, draw=teal, fill=teal, fill opacity=0.2] 
table [x expr=\thisrow{br},y expr=\thisrow{nr}] \mydata;
\addplot[ybar interval, draw=red, fill=red, fill opacity=0.3] 
table [x expr=\thisrow{ba},y expr=\thisrow{na}] \mydata;
\addplot[ybar interval, draw=blue, fill=blue, fill opacity=0.3] 
table [x expr=\thisrow{bn},y expr=\thisrow{nn}] \mydata;
\end{axis}
\end{tikzpicture}
\caption{\footnotesize Histograms of log perplexity evaluated by GPT-2 on real data and collisions.}
\label{fig:lmll}
\end{figure*}

\begin{table*}[t!]
\centering
\footnotesize
\begin{tabular}{l|rr|rr|rr|rr|rr}
\toprule
$\vc$ type & \multicolumn{2}{c|}{MRPC} & \multicolumn{2}{c|}{QQP} & \multicolumn{2}{c|}{Core17/18} & \multicolumn{2}{c|}{Chat} & \multicolumn{2}{c}{CNNDM} \\
 & FP@90 & FP@80 & FP@90 & FP@80  & FP@90 & FP@80  & FP@90 & FP@80  & FP@90 & FP@80  \\  
\midrule
HF & 2.1\% & 0.8\% & 3.1\% & 1.2\% & 4.6\% & 1.2\% & 1.5\% & 0.8\% & 3.2\% & 3.1\% \\
Aggr. & 0.0\% & 0.0\% & 0.0\% & 0.0\% & 0.8\% & 0.7\% & 5.2\% & 2.6\% & 3.1\% & 3.1\% \\
Aggr. $\Omega$ & 47.5\% &  35.6\% & 15.8\% & 11.9\% & 29.3\% & 17.8\% & 76.5\% & 65.3\% & 52.8\% & 35.7\%\\
Nat. & 94.9\% & 89.2\% & 20.5\% & 12.1\% & 13.7\% & 10.9\% & 93.8\% & 86.5\% & 59.8\% & 37.7\% \\
\bottomrule
\end{tabular}
\caption{\footnotesize Effectiveness of perplexity-based filtering.  FP@90 and FP@80 are
false positive rates (percentage of real data mistakenly filtered out)
at thresholds that filter out 90\% and 80\% of collisions, respectively.}
\label{tab:mitigation}
\end{table*}

\subsection{Evaluating Unrelatedness }

We use \textsc{BERTScore}~\cite{zhang2020bertscore} to demonstrate that
our collisions are unrelated to the target inputs.  Instead of exact
matches in raw texts, \textsc{BERTScore} computes a semantic similarity
score, ranging from -1 to 1, between a candidate and a reference by
using contextualized representation for each token in the candidate
and reference.

The baseline for comparisons is \textsc{BERTScore} between the target
input and the ground truth.  For MRPC and QQP, we use $\vx$ as reference;
the ground truth is paraphrases as given.  For Core17/18, we use $\vx$
concatenated with the top sentences except the one with the highest $\gS$
as reference; the ground truth is the sentence in the corpus with the
highest $\gS$.  For Chat, we use the dialogue contexts as reference and
the labeled response as the ground truth.  For CNNDM, we use labeled
summarizing sentences in articles as reference and the given abstractive
summarization as the ground truth.

For MPRC, QQP and CNNDM, we report $F_\textrm{BERT}$ (F$_1$) score.  For
Core17/18 and Chat, we report $P_\textrm{BERT}$ (content from reference
found in candidate) because the references are longer and not token-wise
equivalent to collisions or ground truth.  Table~\ref{tab:bertscore}
shows the results.  The scores for collisions are all negative while the
scores for target inputs are positive, indicating that our collisions
are unrelated to the target inputs.  Since aggressive and regularized
collisions are nonsensical, their contextualized representations are
less similar to the reference texts than natural collisions.

\begin{table}[t]
\centering
\footnotesize
\setlength{\tabcolsep}{5pt}
\begin{tabular}{l|rr|rr}
\toprule
$\vc$ type & \multicolumn{2}{c|}{MRPC} & \multicolumn{2}{c}{Chat} \\
 & BERT & RoBERTa & Bi  $\rightarrow$ Poly & Poly  $\rightarrow$ Bi  \\
\midrule
HF  & 34.0\% & 0.0\% & 55.3\% &  48.9\% \\
Aggr. & 64.5\% & 0.0\% & 77.4\% & 71.3\% \\
Aggr. $\Omega$ & 38.9\% & 0.0\% & 60.5\% &  56.0\%  \\
Nat.  & 41.4\% & 0.0\% & 71.4\% & 68.2\% \\
\bottomrule
\end{tabular}
\caption{\footnotesize Percentage of successfully transferred collisions for MRPC
and Chat.}
\label{tab:transfer}
\end{table}

\subsection{Transferability of Collisions} 

To evaluate whether collisions generated for one target model $f$ are
effective against a different model $f^\prime$, we use MRPC and Chat
datasets.  For MRPC, we set $f^\prime$ to a BERT base model trained with a
different random seed and a RoBERTa model.  For Chat, we use Poly-encoder
as $f^\prime$ for Bi-encoder $f$, and vice versa.  Both Poly-encoder and
Bi-encoder are fine-tuned from the same pretrained transformer model.
We report the percentage of successfully transferred attacks, e.g.,
$\gS(\vx, \vc) > 0.5$ for MRPC and $r=1$ for Chat.

Table~\ref{tab:transfer} summarizes the results.  All collisions achieve
some transferability (40\% to 70\%) if the model architecture is the
same and $f,f^\prime$ are fine-tuned from the same pretrained model.
Furthermore, our attacks produce more transferable collisions than the
HotFlip baseline.  No attacks transfer if $f,f^\prime$ are fine-tuned
from different pretrained models (BERT and RoBERTa).  We leave a study
of transferability of collisions across different types of pretrained
models to future work.


\section{Mitigation}

\noindent
\textbf{\em Perplexity-based filtering.}
Because our collisions are synthetic rather than human-generated texts,
it is possible that their perplexity under a language model (LM) is
higher than that of real text.  Therefore, one plausible mitigation is
to filter out collisions by setting a threshold on LM perplexity.

Figure~\ref{fig:lmll} shows perplexity measured using
GPT-2~\cite{radford2019language} for real data and collisions for each
of our attacks.  We observe a gap between the distributions of real
data and aggressive collisions, showing that it might be possible to
find a threshold that discards aggressive collisions while retaining
the bulk of the real data.  On the other hand, constrained collisions
(regularized or natural) overlap with the real data.


We quantitatively measure the effectiveness of perplexity-based filtering
using thresholds that would discard 80\% and 90\% of collisions,
respectively.  Table~\ref{tab:mitigation} shows the false positive rate,
i.e., fraction of the real data that would be mistakenly filtered out.
Both HotFlip and aggressive collisions can be filtered out with little
to no false positives since both are nonsensical.  For regularized or
natural collisions, a substantial fraction of the real data would be lost,
while 10\% or 20\% of collisions evade filtering.  On MRPC and Chat,
perplexity-based filtering is least effective, discarding around 85\%
to 90\% of the real data.


\paragraphbe{Learning-based filtering.}
Recent works explored automatic detection of generated texts using
a binary classifier trained on human-written and machine-generated
data~\cite{zellers2019defending, ippolito2020automatic}.  These
classifiers might be able to filter out our collisions\textemdash assuming
that the adversary is not aware of the defense.

As a general evaluation principle~\cite{carlini2019evaluating},
any defense mechanism should assume that the adversary has complete
knowledge of how the defense works.  In our case, a stronger adversary
may use the detection model to craft collisions to evade the filtering.
We leave a thorough evaluation of these defenses to future work.

\paragraphbe{Adversarial training.}
Including adversarial examples during training can be effective against
inference-time attacks~\cite{madry2018towards}.  Similarly, training
with collisions might increase models' robustness against collisions.
Generating collisions for each training example in each epoch can be very
inefficient, however, because it requires additional search on top of
gradient optimization.  We leave adversarial training to future work.

\section{Conclusion}

We demonstrated a new class of vulnerabilities in NLP applications:
semantic collisions, i.e., input pairs that are unrelated to each
other but perceived by the application as semantically similar.
We developed gradient-based search algorithms for generating collisions
and showed how to incorporate constraints that help generate more
``natural'' collisions.  We evaluated the effectiveness of our attacks
on state-of-the-art models for paraphrase identification, document and
sentence retrieval, and extractive summarization.  We also demonstrated
that simple perplexity-based filtering is not sufficient to mitigate
our attacks, motivating future research on more effective defenses.

\paragraphbe{Acknowledgements.}
This research was supported in part by NSF grants 1916717 and 2037519,
the generosity of Eric and Wendy Schmidt by recommendation of the Schmidt
Futures program, and a Google Faculty Research Award.

\bibliography{citation_clean}
\bibliographystyle{acl_natbib}

\appendix

\section{Additional Experiment Details}
\label{sec:appendix}

\begin{table}[h]
\footnotesize
\centering
\begin{tabular}{l|rrrrrrr}
\toprule
\textbf{MRPC}  & $B$ & $K$ & $N$ & $T$ & $\eta$ & $\tau$ & $\beta$ \\
\midrule
Aggr.  & 10 & 30 & 30 & 20 & 0.001 & 1.0 & 0.0 \\
Aggr. $\Omega$ & 5 & 15 & 30 & 30 & 0.001 & 1.0 & 0.8 \\
Nat. & 10 & 128 & 5 & 25 & 0.001 & 0.1 & 0.05 \\ 
\midrule
\addlinespace[1.0em]
\textbf{QQP} & \\
\midrule
Aggr.  &  10 & 30 & 30 & 15 & 0.001 & 1.0 & 0.0\\
Aggr. $\Omega$ & 5 & 15 & 30 & 30 & 0.001 & 1.0 & 0.8 \\
Nat. & 10 & 64 & 5 & 20 & 0.001 & 0.1 & 0.0 \\
\midrule
\addlinespace[1.0em]
\textbf{Core} & \\
\midrule
Aggr. & 5 & 50 & 30 & 30 & 0.001 & 1.0 & 0.0 \\
Aggr. $\Omega$ & 5 & 40 & 30 & 60 & 0.001 & 1.0 & 0.85 \\
Nat. & 10 & 150 & 5 & 35 & 0.001 & 0.1 & 0.015 \\
\midrule
\addlinespace[1.0em]
\textbf{Chat} & \\
\midrule
Aggr.  & 5 & 30 & 30 & 15 & 0.001 & 1.0 & 0.0 \\
Aggr. $\Omega$ & 5 & 20 & 30 & 25 & 0.001 & 1.0 & 0.8 \\
Nat. & 10 & 128 & 5 & 20 & 0.001 & 0.1 & 0.15 \\
\midrule
\addlinespace[1.0em]
\textbf{Summ} & \\
\midrule
Aggr. & 5 & 10 & 30 & 15 & 0.001 & 1.0 & 0.0\\
Aggr. $\Omega$ & 5 & 10 & 30 & 30 & 0.001 & 1.0 & 0.8 \\
Nat. & 5 & 64 & 5 & 20 & 0.001 & 1.0 & 0.02 \\
\bottomrule
\end{tabular}
\caption{\footnotesize Hyper-parameters for each experiment. $B$ is the beam size for beam search. $K$ is the number of top words evaluated at each optimization step. $N$ is the number of optimization iterations. $T$ is the sequence length. $\eta$ is the step size for optimization. $\tau$ is the temperature for softmax. $\beta$ is the interpolation parameter in equation \ref{eq:local}.}
\label{tab:params}
\end{table}

\paragraphbe{Hyper-parameters.}
We report the hyper-parameter values for our experiments in
Table~\ref{tab:params}.  The label-smoothing parameter $\epsilon$
for aggressive collisions is set to 0.1.  The hyper-parameters for the
baseline are the same as for aggressive collisions.


\paragraphbe{Runtime.}
On a single GeForce RTX 2080 GPU, our attacks generate collisions in 10 to 60 seconds depending on the length of target inputs.



\newpage 
\section{Additional Collision Examples}

Tables~\ref{tab:para_example},~\ref{tab:doc_example},~\ref{tab:chat_example},~\ref{tab:summ_example}
show collision additional examples for MRPC/QQP,  Core17/18, Chat, and CNNDM respectively.

\label{sec:example}
\begin{table}[h!]
\centering
\footnotesize
\begin{tabularx}{0.48\textwidth}{>{\raggedright}X|c}
\toprule
MRPC/QQP target inputs and collisions & Outputs\\
\midrule
{\bf  \em MRPC Input $(\vx)$}: PCCW 's chief operating officer, Mike Butcher, and Alex Arena, the chief financial officer, will report directly to Mr So.  & \\
{\bf \em Aggressive $(\vc)$}: primera metaphysical declaration dung southernmost among structurally favorably endeavor from superior morphology indirectly materialized yesterday sorority would indirectly ⟨ sg & 99.5\%\\ 
{\bf \em Regularized aggressive $(\vc)$}: in one time rave rave — in … " in but … rv rv smacked a a of a a a a a a a a a of a a & 81.6\% \\ 
{\bf \em Natural $(\vc)$}: in 1989 and joined the new york giants in 1990 & 81.7\%\\ 
\midrule
{\bf  \em MRPC Input $(\vx)$}: Under terms of the deal, Legato stockholders will receive 0.9 of a share of EMC common stock for each share of Legato stock.  & \\
{\bf \em Aggressive $(\vc)$}: moreover author elk telling assert honest exact inventions locally mythical confirms newer feat said assert according locally prefecture municipal realization & 96.7\%\\ 
{\bf \em Regularized aggressive $(\vc)$}: in new " news lust release " on connected different " vibe " reassure females and and to to and and to and to and to and to and to & 95.0\%\\ 
{\bf \em Natural $(\vc)$}: she is also a member of the united states house of representatives, serving as a representative & 83.4\% \\ 
\midrule
{\bf  \em QQP Input $(\vx)$}: How can I slowly lose weight?  & \\
{\bf \em Aggressive $(\vc)$}: sustain fitness recover bru become bolst Enhanced additional distinguished contend crunch Cutting Vital Time cov & 80.5\%\\ 
{\bf \em Regularized aggressive $(\vc)$}: fat Ensure burner www Enhancement Lar Cure Dou St Reaper of of of of of a to and to the the the and to to to of of a of & 85.2\%\\ 
{\bf \em Natural $(\vc)$}: be able that in less long time it & 80.2\% \\ 
\bottomrule
\end{tabularx}
\caption{\footnotesize Collision examples for MRPC and QQP.  Outputs are the probability
scores produced by the model for whether the input and the collisions
are paraphrases.}
\label{tab:para_example}
\end{table}

\makeatletter
\setlength{\@fptop}{0pt}
\makeatother

\begin{table}[t!]
\centering
\footnotesize
\begin{tabularx}{0.48\textwidth}{>{\raggedright}X|c}
\toprule
Core17/18 query inputs and collisions & $r$ \\
\midrule
{\bf  \em Query $(\vx)$}: abuses of e-mail  & \\
{\bf \em Aggressive $(\vc)$}: trailing helsinki, competent regimes internally outlaw wireless offence road : cables by nhs sided head lockheed ford announce oblast million offenders climb ranged postal courier administrations courtesy guangdong oracle & 1 \\
{\bf \em Regularized aggressive $(\vc)$}: un / australia overthrow " — of most telegraph telegraph operations " : the state office in consensus in document lifts down us " by trial " for using < the a and a to and a and a to the a to a a to to a a and a a and a a a the a to to & 1\\ 
{\bf \em Natural $(\vc)$}: the itc ordered all wireless posts confiscated and usps were stripped of their offices and property, leading to a number of & 3 \\
\midrule
{\bf  \em Query $(\vx)$}: heroic acts  & \\
{\bf \em Aggressive $(\vc)$}: colossal helmet vedic bro axes resembling neighbours lead floods blacksmith : evening eligibility caller indicates sculptor coroner lakshmi' than lama announced seizure branded, crafts informing nottinghamshire watch commission. & 1\\ 
{\bf \em Regularized aggressive $(\vc)$}: recorded health and human execution followed, applause prompted, support increased extended : thayer and some there danger, while frank teammate followed feat of personal injury injuries of a the a of the a of the the of of the and of of of of and of of of of and of and of of of the & 1\\ 
{\bf \em Natural $(\vc)$}: the american fighter ( 1 november 1863 ; kia for his feat ) — the japanese ship carrying the cargo of wheat from australia to sydney & 11\\ 
\midrule
{\bf  \em Query $(\vx)$}: cult lifestyles  & \\
{\bf \em Aggressive $(\vc)$}: indiana - semiconductor cut and radiating fire damage, domain reproductive nighttime pastoral calendar failing critical soils indicates force practice ritual belarus stall ; cochin sabha fragmented nut dominance owing & 1\\ 
{\bf \em Regularized aggressive $(\vc)$}: preferred fruits, as willow, suggested to botanist ro spike'for resident nursery : big spreads of pipe rolls and other european pie, a long season at the a and a a and the and of of and of the a of and of of and of of and of of of of and of of the & 2\\ 
{\bf \em Natural $(\vc)$}: the early 1980s their appeal soared : during summerslam in los angeles ( 1993 ), a large number of teenagers went to church to confess their connection to the & 15\\ 
\midrule
{\bf  \em Query $(\vx)$}: art, stolen, forged  & \\
{\bf \em Aggressive $(\vc)$}: colossal helmet vedic bro axes resembling neighbours lead floods blacksmith : evening eligibility caller indicates sculptor coroner lakshmi'than lama announced seizure branded, crafts informing nottinghamshire watch commission & 1\\ 
{\bf \em Regularized aggressive $(\vc)$}: - house and later car dead with prosecutors remaining : “ and cathedral gallery ’ import found won british arrest prosecution a a portrait or mural ( patron at from the the to the a and a to the a and to the a to the of a and to the the and to the to the a and a
 & 3\\ 
{\bf \em Natural $(\vc)$}: the work which left its owner by a mishandle - the royal academy's chief judge inquest  & 8\\ 
\bottomrule
\end{tabularx}
\caption{\footnotesize Collision examples for Core17/18. $r$  are the ranks of irrelevant articles after inserting the collisions.}
\label{tab:doc_example}
\end{table}

\begin{table}[t!]
\centering
\footnotesize
\begin{tabularx}{0.48\textwidth}{>{\raggedright}X|c}
\toprule
Chat target inputs and collisions & $r$\\
\midrule
{\bf  \em Context $(\vx)$}: i'm 33 and love giving back i grew up poor. i did too , back during the great depression. &  \\
{\bf \em Aggressive $(\vc)$}: that to existed with and that is with cope warlord s s came the on  & 1\\
{\bf \em Regularized aggressive $(\vc)$}: camps wii also until neutral in later addiction and the the the the of to and the the the of to and to the the  & 1 \\ 
{\bf \em Natural $(\vc)$}: was the same side of abject warfare that had followed then for most people in this long  & 1  \\ 
\midrule
{\bf  \em Context $(\vx)$}: i am a male . i have a children and a dogs . hey there how is it going ? &  \\
{\bf \em Aggressive $(\vc)$}: is frantically in to it programs s junior falls of it s talking a juan  & 1\\
{\bf \em Regularized aggressive $(\vc)$}: in on from the it department with gabrielle and the the and a and a a to a a and of and of and of  & 1 \\ 
{\bf \em Natural $(\vc)$}: as of this point, and in the meantime it's having very technical support : it employs  & 1  \\ 
\midrule
{\bf  \em Context $(\vx)$}: hi ! how are you doing today ? great , just ate pizza my favorite . . and you ? that's not as good as shawarma &  \\
{\bf \em Aggressive $(\vc)$}: safer to eat that and was mickey in a cut too on it s foreigner & 1\\
{\bf \em Regularized aggressive $(\vc)$}: dipped in in kai tak instead of that and the the a of a of a to to the to and a a of a  & 1 \\ 
{\bf \em Natural $(\vc)$}: not as impressive, its artistic production provided an environment  & 1  \\ 
\bottomrule
\end{tabularx}
\caption{\footnotesize Collision examples for Chat. $r$ are the ranks of collisions
among the candidate responses.}
\label{tab:chat_example}
\end{table}

\begin{table}[t]
\centering
\footnotesize
\begin{tabularx}{0.48\textwidth}{>{\raggedright}X|c}
\toprule
CNNDM ground truth and collisions & $r$\\
\midrule
{\bf  \em Truth}: zayn malik is leaving one direction . rumors about such a move had started since malik left the band 's tour last week . &  \\
{\bf \em Aggressive $(\vc)$}: bp interest yd £ offering funded fit literacy 2020 can propose amir pau laureate conservation  & 1\\
{\bf \em Regularized aggressive $(\vc)$}: the are shortlisted to compete 14 times zealand in in the 2015 zealand artist yo a to to to to to to to to to to to to to to  & 1 \\ 
{\bf \em Natural $(\vc)$}: an estimated \$2 billion by 2014 ; however estimates suggest only around 20 percent are being funded from  & 1  \\ 
\midrule
{\bf  \em Truth}: she says sometimes his attacks are so violent, she's had to call the police to come and save her. &  \\
{\bf \em Aggressive $(\vc)$}: bwf special editor councils want qc iec melinda rey marry selma iec qc disease translated  & 1\\
{\bf \em Regularized aggressive $(\vc)$}: poll is in 2012 eight percent b dj dj dj coco behaviors in dj coco and a a to of to to to the a a to the to a  & 1 \\ 
{\bf \em Natural $(\vc)$}: first national strike since world war ii occurred between january 13 – 15 2014 ; this date will occur  & 1  \\ 
\bottomrule
\end{tabularx}
\caption{\footnotesize Collision examples for CNNDM.  \textbf{\em Truth} are the
true summarizing sentences. $r$ are the ranks of collisions among all
sentences in the news articles.}
\label{tab:summ_example}
\end{table}

\end{document}